\documentclass[10pt,conference]{IEEEtran}
\usepackage{cite}
\usepackage{amsmath,amssymb,amsfonts}
\usepackage{algorithmic}
\usepackage{graphicx}
\usepackage{textcomp}
\usepackage{xcolor}
\usepackage{booktabs}
\usepackage{multirow}
\usepackage{subcaption}
\usepackage{hyperref}
\usepackage{url}
\usepackage{listings}

\usepackage{float}

\definecolor{codegreen}{rgb}{0,0.6,0}
\definecolor{codegray}{rgb}{0.5,0.5,0.5}
\definecolor{codepurple}{rgb}{0.58,0,0.82}
\definecolor{backcolour}{rgb}{0.95,0.95,0.95}

\lstdefinestyle{mystyle}{
    backgroundcolor=\color{backcolour},   
    commentstyle=\color{codegreen},
    keywordstyle=\color{magenta},
    numberstyle=\tiny\color{codegray},
    stringstyle=\color{codepurple},
    basicstyle=\ttfamily\footnotesize,
    breakatwhitespace=false,         
    breaklines=true,                 
    captionpos=b,                    
    keepspaces=true,                 
    numbers=left,                    
    numbersep=5pt,                  
    showspaces=false,                
    showstringspaces=false,
    showtabs=false,                  
    tabsize=2
}

\begin{document}

\title{Multi-Modal Vision vs. Text-Based Parsing: Benchmarking LLM Strategies for Invoice Processing}

\author{\IEEEauthorblockN{
David Berghaus\IEEEauthorrefmark{2}\IEEEauthorrefmark{3}, Armin Berger\IEEEauthorrefmark{2}, Lars Hillebrand\IEEEauthorrefmark{2}, Kostadin Cvejoski\IEEEauthorrefmark{2}\IEEEauthorrefmark{3},
Rafet Sifa\IEEEauthorrefmark{2}} \\
\IEEEauthorrefmark{2}\textit{Fraunhofer IAIS}, Sankt Augustin, Germany \\
\IEEEauthorrefmark{3}\textit{Lamarr Institute}, Germany
}

\maketitle

\begin{abstract}
This paper benchmarks eight multi-modal large language models from three families (GPT-5, Gemini 2.5, and open-source Gemma 3) on three diverse openly available invoice document datasets using zero-shot prompting. We compare two processing strategies: direct image processing using multi-modal capabilities and a structured parsing approach converting documents to markdown first. Results show native image processing generally outperforms structured approaches, with performance varying across model types and document characteristics. This benchmark provides insights for selecting appropriate models and processing strategies for automated document systems. Our code is available online.\footnote{\url{https://anonymous.4open.science/r/invoice_benchmark_paper-4361/README.md}}
\end{abstract}


\section{Introduction}
Invoice and order processing represents a ubiquitous and resource-intensive business function across industries. For decades, enterprises had to either use manual human labour for these tedious tasks or specialized Optical Character Recognition (OCR) and template-based systems to automate these workflows, which required extensive customization, struggled with document variability, and demanded continuous maintenance as document formats evolve.

The emergence of multi-modal large language models (LLMs) has created a paradigm shift in document understanding capabilities. These models can simultaneously process visual information, recognize text layout, infer semantic relationships, and extract structured data. Models such as OpenAI's GPT-5, Google's Gemini 2.5, and open-source alternatives like Google's Gemma represent the cutting edge of this technological advancement, offering potential for more generalizable and adaptable document processing solutions.

Despite the widespread interest in deploying these technologies for financial document automation, there has been limited systematic evaluation of their performance on real-world invoice processing tasks. Moreover, given the often challenging layouts of invoices, as well as potentially noisy scans, it is unclear how the performance of multi-modal vision approaches compare to more traditional \emph{parsing to text} strategies. 

This research addresses this critical need by providing a comprehensive benchmark framework for evaluating different LLM approaches for invoice processing. Our study makes the following key contributions:

\begin{itemize}
    \item We introduce a unified benchmarking methodology that enables direct comparison of document processing performance across different model families, document types, and extraction fields
    
    \item We evaluate eight state-of-the-art models from three major families (OpenAI, Google Gemini, and Google Gemma) on three diverse document datasets representing different quality levels and document types
    
    \item We compare two distinct processing strategies: direct image processing that leverages the models' multi-modal capabilities, and a structured parsing approach that first converts document images to markdown using Docling before LLM processing
    
    \item We provide detailed analysis of performance variations across different document sections (headers, line items, summaries) and field types (numeric, textual, unstructured identifiers like IBANs)
    
    \item We examine the implications of model size and efficiency, comparing smaller variants (gpt-5-nano, gemini-2.5-flash-lite, gemma-3-4b-it) against their larger counterparts
\end{itemize}

The insights from this benchmark can help organizations make informed decisions when selecting models and processing strategies for their document automation needs. Furthermore, our analysis highlights current limitations and areas for improvement in multi-modal document processing capabilities, providing direction for future research and development efforts in this rapidly evolving field.

\section{Related Work}
\label{sec:related_work}
LLMs have been used for business workflow automation in various contexts \cite{pfeiffer2025theorypracticerealworlduse,10.1145/3342558.3345421,10825431,10825159}
Specifically for invoice analysis, the survey article \cite{10418145} lists recent progress of ML approaches in invoice data extraction techniques but does not cover multi-modal LLMs. 

The evolution of document understanding from traditional OCR to modern multi-modal approaches has been well-documented. \cite{Subramani2020ASO} provide a foundational survey categorizing deep learning methods, while \cite{ding2024deeplearningbasedvisually} offer a specialized analysis of multi-modal techniques for visually rich documents, highlighting the importance of joint vision-text representations for layout understanding.

\cite{liu2024memoryaugmentedagenttrainingbusiness} demonstrate how memory-augmented LLM agents can iteratively refine extraction capabilities through experience, showing particular promise for logistics invoice processing where transport reference identification requires contextual understanding of document structures.

Benchmarking these systems presents unique challenges. While several benchmarks for multi-modal models and document understanding exist \cite{wornow2024wonderbreadbenchmarkevaluatingmultimodal,ma2024mmlongbenchdoc,chia2024mlongdocbenchmarkmultimodalsuperlong,zou2024docbenchbenchmarkevaluatingllmbased}, to the best of our knowledge, results on the performance of LLMs for invoice processing tasks have so far been missing.

\section{Methodology}
\label{sec:methodology}

\subsection{Benchmark Framework}
We developed a comprehensive unified benchmark framework to systematically evaluate different models and processing strategies across multiple datasets. Each model's responses are compared against ground truth annotations with appropriate normalization to handle variations in formatting (such as different date formats or whitespace variations).

\subsection{Models}
We evaluated eight state-of-the-art multi-modal models across three families, representing both commercial and open-source offerings. Our selection includes three models from the OpenAI GPT 5 family (gpt-5-chat, gpt-5-mini, and gpt-5-nano), three from the Google Gemini 2.5 family (gemini-2.5-pro, gemini-2.5-flash, and gemini-2.5-flash-lite), and two open-source models from the Google Gemma 3 family (gemma-3-12b-it and gemma-3-4b-it \cite{gemma3}).

\subsection{Processing Strategies}
We compared two distinct processing strategies that represent fundamentally different approaches to document understanding:

\begin{itemize}
    \item \textbf{Native Image Processing}: In this approach, document images are provided directly to multi-modal LLMs, which analyze the visual content and extract structured information. This leverages the models' abilities to understand document layout, recognize text visually, and interpret spatial relationships between elements. The advantage of this approach is that it preserves all visual information and spatial context present in the original document.
    
    \item \textbf{Docling Processing}: This two-step approach first uses Docling \cite{Docling} (we also tested with SmolDocling \cite{SmolDocling} but the results were almost identical so we do not report them here) which is a wide-spread open-source tool with SOTA performance \cite{docling_benchmark} that analyzes document layout, recognizes text elements, and renders them in a markdown format that preserves structural information while standardizing the presentation. This conversion process creates a text-only representation that maintains tables, sections, and other layout elements through markdown syntax.
    
    This approach potentially reduces the visual complexity of the task for LLMs and provides a more structured input, but may lose some visual context from the original document.
\end{itemize}

For both strategies, we used carefully designed prompts tailored to each dataset's specific structure and extraction requirements. The prompts were designed to elicit structured JSON outputs that could be directly compared with ground truth annotations. The same prompt templates were used across all models within each processing strategy to ensure fair comparison.

\subsection{Datasets}
Our benchmark includes three diverse datasets representing different document types, quality levels, and complexities:

\begin{enumerate}
    \item \textbf{Clean Invoices (Donut)}: A dataset of 500 synthetic invoices \cite{donut} with structured JSON annotations for document understanding tasks. The collection features diverse invoice layouts containing key business information (seller/client details, line items with quantities/prices, and financial totals).
    
    \item  \textbf{Scanned Receipts (ICDAR-2019-SROIE)}: A dataset of 1000 scanned receipt images \cite{icdar} collected for the ICDAR 2019 competition on scanned receipt OCR and Information Extraction. The dataset includes receipts exhibiting a wide range of real-world variations including differing layouts, print quality, and image conditions (e.g., skew, lighting). Annotations include bounding boxes for text localization, OCR transcripts, and key information fields (company, address, date, and total amount) encoded in JSON. 
    
    \item \textbf{Scanned Invoices (inv-cdip)}: A dataset of 350 annotated invoice images derived by \cite{inv-cdip} from the Tobacco Collections of Industry Documents Library \cite{ucsf_tobacco_library}, focusing on structured field extraction. The documents exhibit real-world scanning artifacts such as stamps, handwritten annotations, and variable print quality. Each invoice is annotated with 7 key fields: invoice number, purchase order, invoice date, due date, amount due, total amount, and total tax. We remark that we found some inconsistencies in the annotations, namely for one field, different wordings ("total net worth" and "item net worth") were used in 14 samples. We corrected this for the experiments in this work.
\end{enumerate}

All datasets are stored as images, meaning that OCR techniques have to be used by the models in order to convert them to text. It would be interesting to conduct similar experiments with PDF documents with native texts, unfortunately we did not find suitable datasets for this. We also remark that the datasets are open source, so it is possible that they have been used as training data for the large language models.

\subsection{Evaluation Metrics}
We evaluated performance using a comprehensive set of metrics designed to provide insights at multiple levels of granularity:

\begin{itemize}
    \item \textbf{Overall Accuracy}: The percentage of correctly extracted fields across all sections, calculated as the ratio of correctly extracted fields to total fields in the ground truth. This provides a high-level performance indicator across the entire document.
    
    \item \textbf{Field-specific Accuracy}: Accuracy measurements for individual fields, with particular attention to challenging fields like IBAN numbers, entity names, and line item descriptions. This reveals specific strengths and weaknesses across different data types.
\end{itemize}

For all metrics, we considered a field to be correctly extracted if it matched the ground truth exactly after minimal normalization. Normalization included standardizing date formats and normalizing whitespace but required exact matching of the core information content. We did not correct for commas and dots in numerical values since it can be very important for correct order processing.

\begin{table}[htbp]
\centering
\caption{Model Performance Comparison Across Datasets}
\label{tab:model_performance}
\begin{tabular}{lllr}
\toprule
Dataset & Model & Version & Overall Acc. (\%) \\
\midrule
\multicolumn{4}{l}{\textbf{Scanned Receipts}} \\
\midrule
 & gemini-2.5-pro & native & \textbf{87.46\%} \\
 &  & smoldocling & 46.54\% \\
 & gemini-2.5-flash & native & 82.15\% \\
 &  & smoldocling & 47.00\% \\
 & gemini-2.5-flash-lite & native & 85.34\% \\
 &  & smoldocling & 44.69\% \\
 & gpt-5-chat & native & 69.21\% \\
 &  & smoldocling & 42.37\% \\
 & gpt-5-mini & native & 53.92\% \\
 &  & smoldocling & 32.59\% \\
 & gpt-5-nano & native & 58.08\% \\
 &  & smoldocling & 37.54\% \\
 & google/gemma-3-12b-it & native & 66.91\% \\
 &  & smoldocling & 41.48\% \\
 & google/gemma-3-4b-it & native & 53.79\% \\
 &  & smoldocling & 36.69\% \\
\midrule
\multicolumn{4}{l}{\textbf{Clean Invoices}} \\
\midrule
 & gemini-2.5-pro & native & \textbf{96.50\%} \\
 &  & smoldocling & 85.14\% \\
 & gemini-2.5-flash & native & 93.55\% \\
 &  & smoldocling & 84.13\% \\
 & gemini-2.5-flash-lite & native & 95.70\% \\
 &  & smoldocling & 84.88\% \\
 & gpt-5-chat & native & 96.03\% \\
 &  & smoldocling & 84.89\% \\
 & gpt-5-mini & native & 96.31\% \\
 &  & smoldocling & 84.88\% \\
 & gpt-5-nano & native & 86.45\% \\
 &  & smoldocling & 85.45\% \\
 & google/gemma-3-12b-it & native & 84.66\% \\
 &  & smoldocling & 84.91\% \\
 & google/gemma-3-4b-it & native & 45.69\% \\
 &  & smoldocling & 84.59\% \\
\midrule
\multicolumn{4}{l}{\textbf{Scanned Invoices}} \\
\midrule
 & gemini-2.5-pro & native & \textbf{92.71\%} \\
 &  & smoldocling & 63.94\% \\
 & gemini-2.5-flash & native & 92.07\% \\
 &  & smoldocling & 62.70\% \\
 & gemini-2.5-flash-lite & native & 91.64\% \\
 &  & smoldocling & 61.70\% \\
 & gpt-5-chat & native & 88.01\% \\
 &  & smoldocling & 64.03\% \\
 & gpt-5-mini & native & 86.88\% \\
 &  & smoldocling & 55.99\% \\
 & gpt-5-nano & native & 80.61\% \\
 &  & smoldocling & 55.29\% \\
 & google/gemma-3-12b-it & native & 82.09\% \\
 &  & smoldocling & 61.21\% \\
 & google/gemma-3-4b-it & native & 76.02\% \\
 &  & smoldocling & 55.43\% \\
\bottomrule
\end{tabular}
\end{table}

\section{Results}
\label{sec:results}

Our evaluation results, summarized in Table \ref{tab:model_performance}, reveal several key patterns in model performance and the efficacy of different processing strategies.

\subsection{Processing Strategy Comparison}

The most significant finding is the consistent and substantial superiority of native image processing over the structured parsing (Docling) approach. Across all datasets and models, direct image analysis yielded significantly higher accuracy. For instance, on the Scanned Receipts dataset, the top-performing model achieved 87.46\% accuracy with native processing, compared to a maximum of only 47.00\% using the Docling method. A similar gap was observed for Scanned Invoices, where native processing reached 92.71\% versus 64.03\% for the parsed text approach.

Notably, the Docling conversion process appears to create a performance bottleneck, particularly on the Clean Invoices dataset. Here, most models performed within a narrow accuracy band of 84-85\%, effectively neutralizing the advanced capabilities of the larger, more powerful models. This suggests that the initial OCR and markdown conversion, rather than the LLMs' reasoning abilities, becomes the primary limiting factor in the structured parsing pipeline.

\subsection{Model Performance}

Among the evaluated models, the Gemini 2.5 family demonstrated the strongest overall performance. Gemini 2.5 Pro consistently achieved the highest accuracy across all three datasets: 87.46\% on Scanned Receipts, 96.50\% on Clean Invoices, and 92.71\% on Scanned Invoices.

The GPT 5 models proved to be highly competitive, especially on the less noisy Clean Invoices dataset, where both GPT 5 Chat and GPT 5 Mini surpassed 96\% accuracy. While their performance on the scanned datasets was robust, it was generally a few percentage points lower than the leading Gemini models.

The open-source Gemma 3 models showed promising results. The larger `gemma-3-12b-it` model delivered solid performance, particularly with native image processing on Scanned Invoices (82.09\%). However, the smaller `gemma-3-4b-it` model struggled significantly with direct image analysis, especially on the Clean Invoices dataset (45.69\%), indicating a capability threshold below which smaller models are less effective for complex visual extraction tasks.

\subsection{Challenging Fields and Efficiency}

\begin{figure}[H]
    \centering
    \includegraphics[width=1\linewidth]{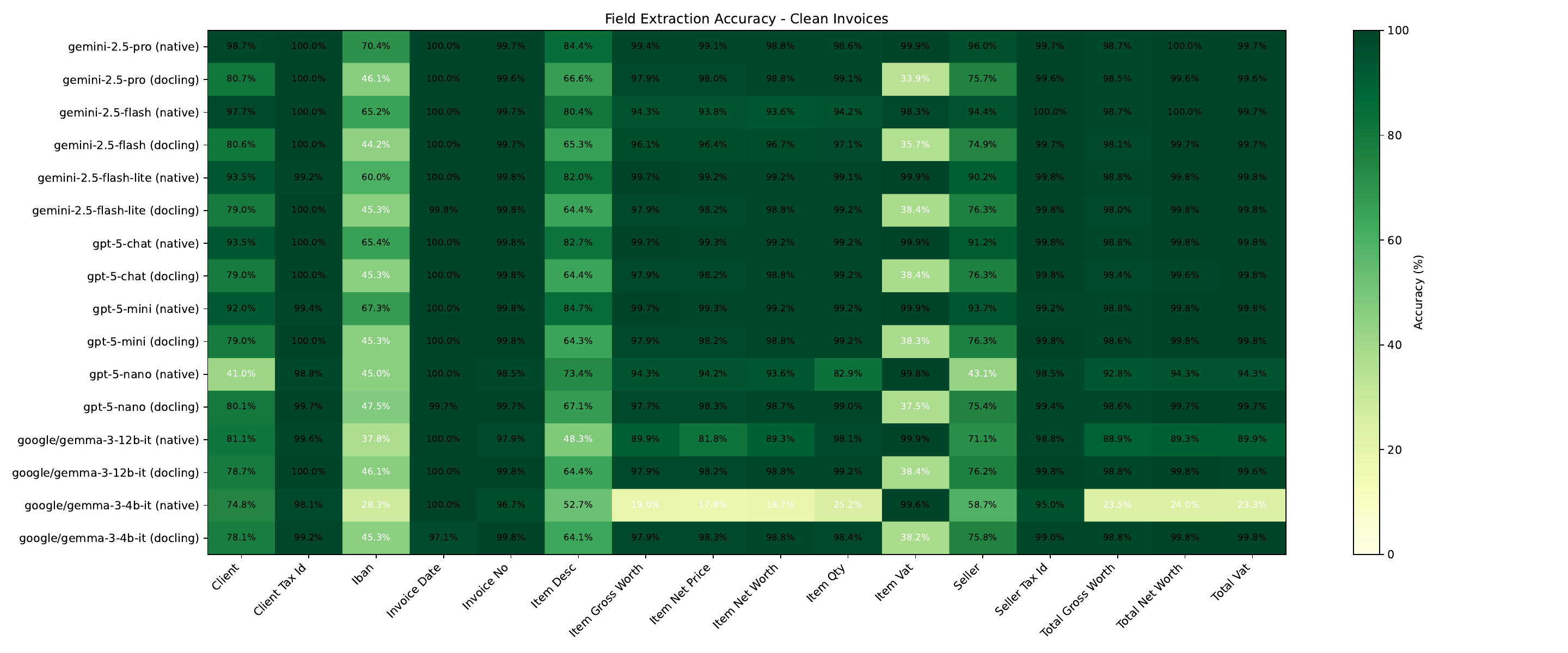}
    \caption{Field extraction performance on the Clean Invoices (Donut \cite{donut}) dataset}
    \label{fig:donut-extraction}
\end{figure}

As becomes evident from figure \ref{fig:donut-extraction}, the extraction performance on the IBAN field for the clean invoices dataset \cite{donut} is significantly worse than for the other fields. Common mistakes appear to be OCR-related, such as confusing the number zero with the letters "O" or "U". This indicates that highly unstructured alphanumeric fields present particular difficulties even for advanced multi-modal models.

\section{Conclusion and Future Work}
\label{sec:conclusion}

Our benchmark demonstrates that current multi-modal LLMs offer powerful capabilities for automating invoice processing, with native image processing consistently outperforming structured conversion approaches. This suggests that visual context and layout understanding are crucial for effective document processing, and that current models effectively leverage this information when processing images directly.

Among the models evaluated, Gemini variants achieved the highest overall accuracy. The open-source Gemma-3-12b-it emerged as a reasonably competitive alternative, particularly for organizations prioritizing cost-effectiveness or privacy. We anticipate that larger open-source models could further narrow the performance gap with proprietary models.

Several limitations warrant future research attention. The persistent difficulty in extracting unstructured fields like IBANs suggests room for improvement in handling alphanumeric patterns. Performance on noisy scanned documents also remains challenging compared to clean digital invoices.

Future work should explore specialized models and investigate fine-tuning for document understanding tasks. Promising candidates could be LayoutLM \cite{10.1145/3394486.3403172,10.1145/3503161.3548112} and LiLT \cite{Wang2022LiLTAS} which are models specialized for layout understanding. We do not benchmark these models in this work because they require fine-tuning and do not work within our unified zero-shot prompting approach.

In conclusion, our benchmark shows that direct image processing with multi-modal LLMs offers a powerful approach for document automation, with continued advancements in both proprietary and open-source models promising increasingly accessible and capable solutions.

\bibliographystyle{IEEEtran}
\bibliography{bib}

\onecolumn

\begin{appendices}

\section{Field Extraction Performance}

\begin{figure}[H]
    \centering
    \includegraphics[width=1\linewidth]{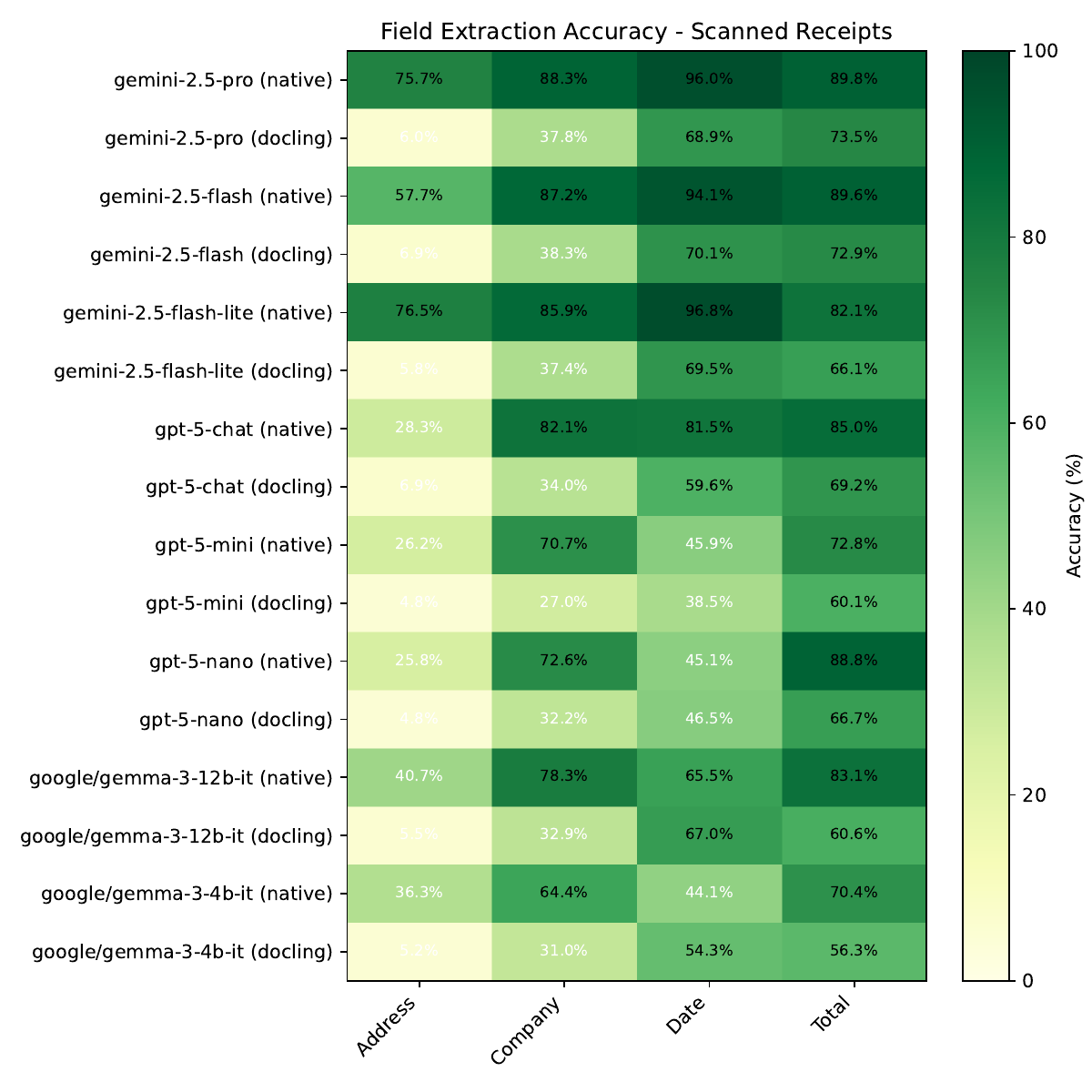}
    \caption{Field extraction performance on the Scanned Receipts (ICDAR \cite{icdar}) dataset}
    \label{fig:icdar-field-extraction}
\end{figure}

\begin{figure}[H]
    \centering
    \includegraphics[width=1\linewidth]{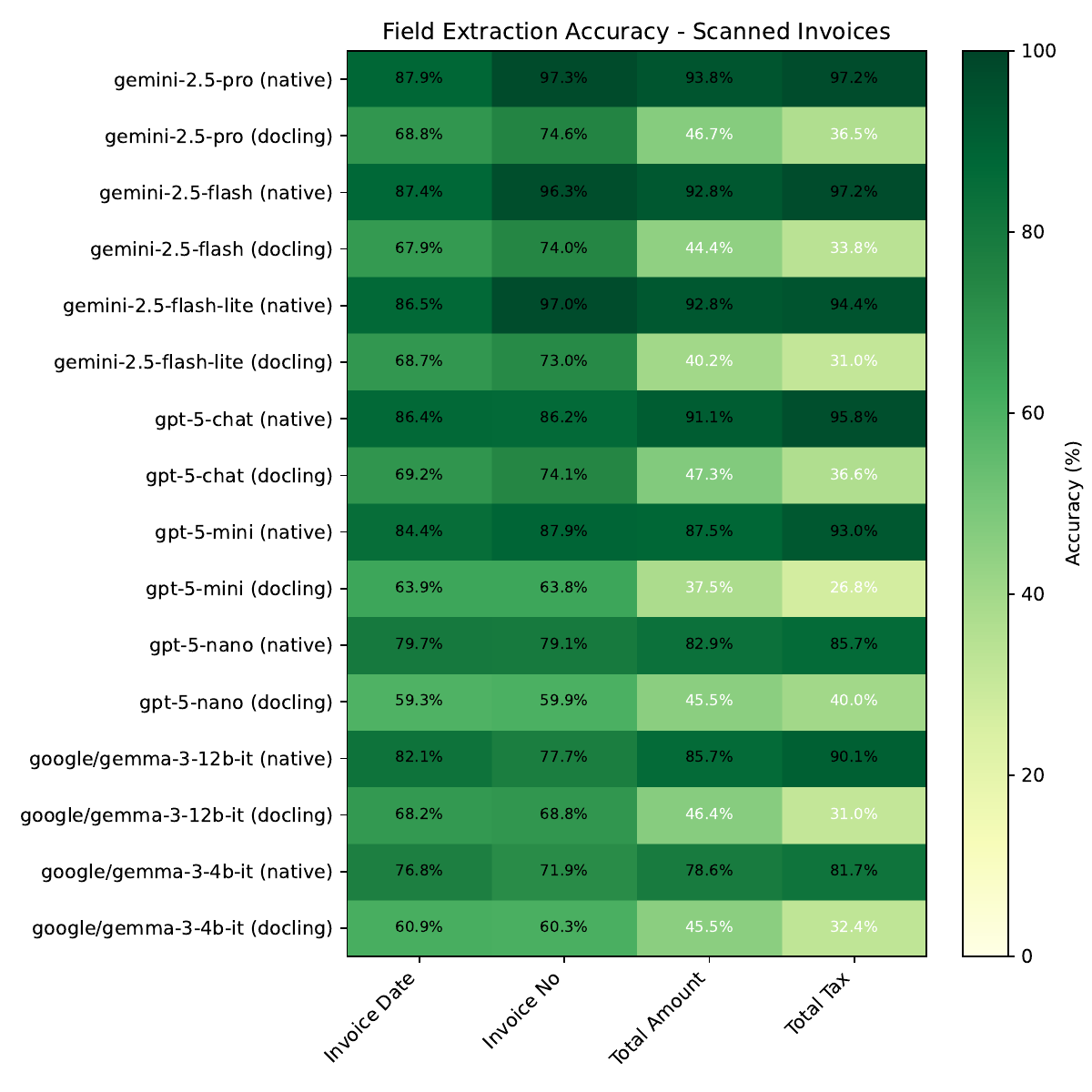}
    \caption{Field extraction performance on the  Scanned Invoices (inv-cdip \cite{inv-cdip}) dataset}
    \label{fig:invcdip-extraction}
\end{figure}

\end{appendices}
\end{document}